\pdfoutput=1
\documentclass[10pt,a4paper]{article}
\usepackage{makecell}
\usepackage{graphicx}
\usepackage[labelfont=bf]{caption}
\usepackage[colorlinks=true, linkcolor=blue, urlcolor=blue, citecolor=blue]{hyperref}
\hypersetup{breaklinks=true}
\usepackage{float}
\usepackage{booktabs}
\usepackage{siunitx}
\usepackage[a4paper]{geometry}
\usepackage[backend=biber, style=nature]{biblatex}

\usepackage[utf8]{inputenc}

\begin{document}

\title{Automated Phenotyping of Epicuticular Waxes of Grapevine Berries Using Light Separation and Convolutional Neural Networks}

\date{}

\maketitle

\author{Pierre Barré$^{1}$, Katja Herzog$^{3}$, Rebecca Höfle$^{3}$, Matthias B. Hullin$^{2}$, Reinhard Töpfer$^{3}$ and Volker Steinhage$^{1}$*}

$^{1}$ \quad University of Bonn, Institute of Computer Science IV, Endenicher Allee 19A, Bonn 53115, Germany \\
$^{2}$ \quad University of Bonn, Institute of Computer Science II, Endenicher Allee 19A, Bonn 53115, Germany\\
$^{3}$ \quad Julius Kühn-Institut(JKI)-Federal Research Centre of Cultivated Plants, Institute for Grapevine Breeding Geilweilerhof, Siebeldingen 76833, Germany\\

{Author to whom correspondence should be addressed: steinhag@cs.uni-bonn.de}

\begin{abstract}{In viticulture the epicuticular wax as the outer layer of the berry skin is known as trait which is correlated to resilience towards \textit{Botrytis} bunch rot. Traditionally this trait is classified using the OIV descriptor 227 (berry bloom) in a time consuming way resulting in subjective and error-prone phenotypic data. In the present study an objective, fast and sensor-based approach was developed to monitor berry bloom. From the technical point-of-view, it is known that the measurement of different illumination components conveys important information about observed object surfaces. A Mobile Light-Separation-Lab is proposed in order to capture illumination-separated images of grapevine berries for phenotyping the distribution of epicuticular waxes (berry bloom). For image analysis, an efficient convolutional neural network approach is used to derive the uniformity and intactness of waxes on berries. Method validation over six grapevine cultivars shows accuracies up to 97.3\%. In addition, electrical impedance of the cuticle and its epicuticular waxes (described as an indicator for the thickness of berry skin and its permeability) was correlated to the detected proportion of waxes with r=0.76. This novel, fast and non-invasive phenotyping approach facilitates enlarged screenings within grapevine breeding material and genetic repositories regarding berry bloom characteristics and its impact on resilience towards \textit{Botrytis} bunch rot.}
\end{abstract}


\newpage
\section{Introduction}
\label{section:section_1_introduction}
\begin{figure}[H]
	\centering
	\includegraphics[scale =0.70]{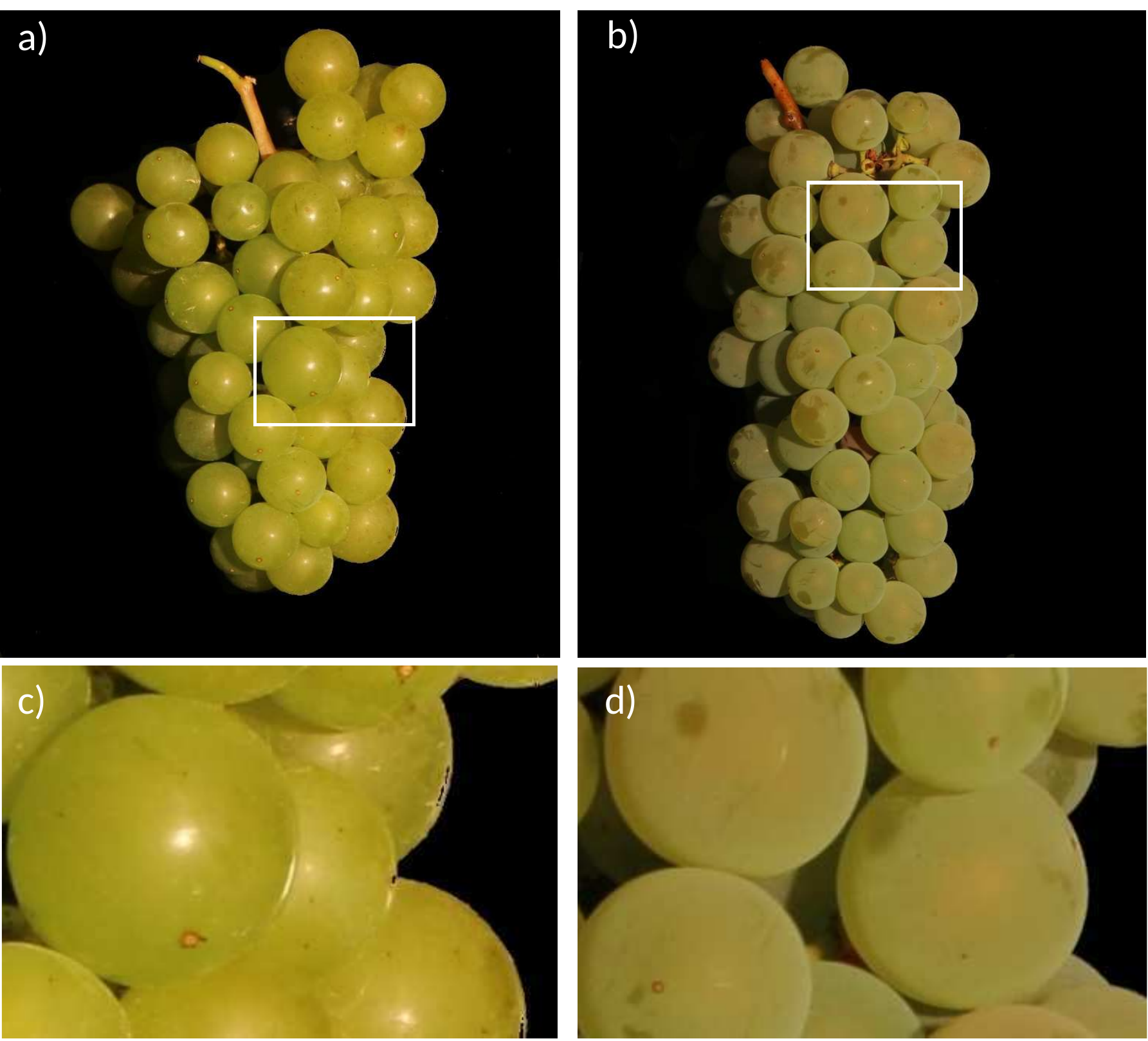}
	\caption{
		Exemplary bunches of the grapevine cultivar Morio Muskat \textbf{(a)} and the breeding strain Seibel 7511 \textbf{(b)}. The berries of Morio Muskat show no epicuticular wax layer \textbf{(c)} and the berries of Seibel 7511 show uniform and intact epicuticular wax layer \textbf{(d)}}
	\label{fig:figure_1}
\end{figure}

The risk for \textit{Botrytis} bunch rot, a disease caused by \textit{Botrytis cinerea} Pers., is seriously increasing in viticultural regions especially when high air humidity or prolonged rain cause in persistent moisture on grapevine berry surfaces. \textit{Botrytis} infestation then reduces grape yield and the quality of wines due to off-flavors or reduced wine stability  in susceptible grapevine cultivars \cite{meneguzzo2008effect}. Besides climatic conditions, susceptibility of grapes against \textit{Botrytis} bunch rot is mainly influenced by morphological properties like the bunch compactness, canopy structure as well as the thickness of the berry skin and hydrophobic characteristics of the berry cuticle (\cite{rosenquist1988development}, \cite{percival1993effect}, \cite{gabler2003correlations}, \cite{molitor2011timing}, \cite{herzog_impedance_2015}). The cuticle and its epicuticular wax layer hereby represents the outer layer of the grapevine berries (cf. figure \ref{fig:figure_1}), it shows semi-crystalline to crystalline structure and influences the retention of pesticides, the hydrophobic characteristics of the grapevine berry surface and the adhesive ability of plant pathogens (citation overview is given by \cite{percival1993effect}). Furthermore, it is described as important economic feature: the berry bloom imparts a frosted appearance to the berry, which is considered attractive and desirable by consumers of table grapes \cite{kok2004determination}. The berry epicuticular waxes influences the oviposition of European grapevine moth \cite{rid2018waxy} but interestingly they are also scattering light. For the grapevine berry cuticle, classification of phenotypes by traditional visual estimations is only possible for the appearance of the wax layer (OIV 227). This method is very time consuming and phenotypic data are subjective with low resolution which renders the differentiation between visibly similar genotypes hardly possible. Thus, more precision phenotyping methods are needed which are often based on labor-intensive light or scanning electron microscopy (\cite{percival1993effect}, \cite{gabler2003correlations}, \cite{becker2012deposition}) as well as expensive and time-consuming chemical analyses (\cite{rid2018waxy}, \cite{ozer2017investigations}).\\
Laborious phenotyping methods are not feasible in grapevine research and breeding where berries of several hundreds of genotypes have to be evaluated in a short time. Regarding objective and high-throughput phenotyping techniques, Herzog et al. \cite{herzog_impedance_2015} developed a sensor-based method in order to characterize the thickness and permeability of grape berry cuticles by measuring the electrical impedance. High impedance is described as an indicator for thick berry cuticle and berry bloom \cite{herzog_impedance_2015}. Within this I-sensor the cuticle and the epicuticular wax layer as well as the epidermal cell layer of berries are located between two electrically conducting compartments: (1) berries placed on a NaCl solution and (2) an electrode within the berry flesh \cite{molitor2018multi}. As described by \cite{molitor2018multi}, the impedance as its own is the sum of the real resistor (permeability of the cuticle, the wax layer and air) and the imaginary resistor (thickness of the cuticle, the wax layer and air). \cite{herzog_impedance_2015} observed a high negative correlation between the grape berry impedance of the cuticule and its epicuticular waxes and the susceptibility of berries to \textit{Botrytis} bunch rot. The measurement of impedance is a point measurement and minimal-invasive due to a spotty puncture. The distribution, the uniformity or the intactness of epicuticular waxes as very important features cannot directly be determined due to point measurements. Subsequent infection tests on the suceptibility towards \textit{Botrytis} bunch rot are not feasible because of the puncture which functions as portal of entry for pathogene.\\
In addition to the non-imaging impedance as indicator for thickness of the grape berry cuticle, imaging methods will enable objective characterization of the appearance and uniformity of berry bloom on the berry surface. Light-separation methods are useful to recognize parts or structures that consumer cameras usually cannot distinguish, e.g. surface texture. \\
From materials science, it is known that different illumination components will be reflected in different ways from the illuminated object surfaces. Therefore, different reflectance components convey important information about the observed object surfaces that cannot be inferred from their sum. Recently, several approaches were published separating illuminations and reflection components (\cite{anderson_polarized_1991}, \cite{wolff1991constraining}, \cite{jacques_imaging_2000}, \cite{jacques_imaging_2002}, \cite{nayar_fast_2006}, \cite{subramaniam_detection_nodate}, \cite{garces_low_2015}). In the present study, we will focus on polarization-based and pattern-based approaches that are most relevant to the scope of this contribution. One key observation is that specular reflections become polarized, i.e., the light waves are aligned in one single plane, while in turn, diffuse reflections are basically unpolarized. Generally, specular reflections are caused by smooth surface patches while diffuse reflections are caused by rough surface patches. Wolff and Boult  \cite{wolff1991constraining} present a polarization reflectance model showing that all polarization-based methods follow from this model. They demonstrate the capability of polarization-based methods to segment material surfaces according to varying levels of relative electrical conductivity, especially by distinguishing non-conducting dielectrics from conductive metals. Nayar et al. \cite{nayar_fast_2006} use controlled illumination to separate direct from global reflections. More precisely, they employ high frequency illumination patterns derived by chessboard-like arrangements of transparent and opaque filter cells. In general, direct reflections are due to direct illumination of surface patches by the light source. Global reflections are due to different physical phenomena including interreflection of light between different scene points but also subsurface scattering within the medium beneath the object surface. Garces et al. \cite{garces_low_2015} present an outdoor upgrade of Nayar’s method \cite{nayar_fast_2006} by employing the sun and sky lighting as incident light sources and creating high frequency patterns by casting a stick over the observed objects. \\
In the given study, we present and evaluate a novel sensor- and computer-based phenotyping method combining the concept of light-separation and deep learning in order to provide an automated phenotyping of epicuticular waxes on the grapevine berry surface, the berry bloom.
We used the polarization-based approach of Wolff and Boult \cite{wolff1991constraining} and the pattern-based approach of Nayar et al. \cite{nayar_fast_2006} to separate illumination and reflections into four channels showing specular reflections, diffuse reflections, direct reflections and global reflections. These four image channels characterize surface patches in terms of smoothness and depth of reflection (surface or sub-surface reflections) and are used to measure epicuticular waxes of grapevine berries. To facilitate light-separation within image capture, we designed and built a mobile device, the LSL (\textbf{L}ight \textbf{S}eparation \textbf{L}ab) for image capture of grapevine berries employing both light-separation methods.\\
While common image analysis approaches to visual detection and measurement of objects include serious feature engineering, i.e. the employment of domain-specific knowledge to create descriptive features for localizing the berries and measuring the epicuticular waxes, deep learning approaches use convolutional neural networks (CNN) that allow for an automated feature derivation. CNNs were successfully applied in plant sciences \cite{imageClef}, especially in order to study plant diseases and in particular leaf lesions. Lu et al. \cite{lu2017field} developed a deep-learning method to detect six crop diseases. Mohanty et al. \cite{mohanty2016using} detect 38 leaf diseases from the PlantVillage data set \cite{hughes2015open} using two approaches: an AlexNet-based CNN \cite{krizhevsky2012imagenet} and the GoogLeNet CNN with the inception architecture \cite{szegedy2015going}. Since they used the same data set, the last work can be directly related with Wang et al. \cite{wang2017automatic} using the concept of transfer learning. Sladojevic et al. \cite{sladojevic2016deep} collected leaf images from the internet which are classified under thirteen kinds of diseases and used a CNN network based on AlexNet \cite{krizhevsky2012imagenet}. \\
The major objective of the present study was the combination and evaluation of light-separation methods with fast and high-resolution learning approaches using Convolutional Neural Networks (CNN) in order to facilitate an automated phenotyping of epicuticular waxes on the grapevine berry surface, the berry bloom. First, we will show the proof-of-concept that the pattern-based and polarization-based light-separation methods improve the accuracy of the detection of berry and its epicuticular waxes. Secondly, we will show the results on precision and validity of our system by computing the proportion of epicuticular waxes on the detected berry surface and the correlation of these phenotypic data to impedance measurements. Thirdly, we will analyze the correlation between impedance measurements on the one hand and the derived surface proportion of the epicuticular waxes for validation.

\section{Results}
\label{section:section_2_results}
In the present study, we employed a workflow of two CNNs: Hereby, standard RGB images and four separated reflection channels were used in the first shifting pattern recognition CNN for detection and localization of berries in images resulting in Areas of Interest (AoI). The same data (RGB and reflection channels) were used in the second CNN in order to segment the detected AoI according to the presence of epicuticular wax. The most important advantage of the developed CNN-based approach is that both CNNs automatically learned to derive appropriate features over the given input channels. We assumed that electrical impedance of the cuticle and its epicuticular wax is related to the presence of the wax layer and would thus be useful as reliable reference parameter. Accordingly, the impedance of each investigated berry (cf. \cite{herzog_impedance_2015}) was measured after the LSL-based image capture.

\subsection{Detection of the Berry and Its Epicuticular Waxes}
\label{subsection:section_2_1_Deep Learning_:_Detection_of_the_Berry_and_Its_Epicuticular_Waxes}
Image capture with the the LSL (Light Separation Lab) generates five outputs: (1) a standard color RGB image without any light separation; (2) a color image showing specular reflections; (3) a color image showing diffuse reflections, (4) a color image showing direct reflections; (5) a color image showing global reflections. 
We evaluated the results of berry detection and quantification of epicuticular waxes on the berry surface with four input modes: (I) only standard color images (1) as baseline for comparison; (II) color images (2) and (3) generated by polarization-based light separation;
(III) color images (4) and (5) generated by pattern-based light separation; (IV) input of all images (1) to (5).
Both CNNs for berry detection and quantification of epicuticular waxes were evaluated using k-fold cross-validation with k = 3 based on 270 berry images captured by the LSL (4.1). In each run of the 3-fold cross-validation 180 images are used for training and 90 images are used for evaluation. More precisely, about 18,000,000 labeled pixels are extracted using a semi-automated annotation tool that allows labeling all pixels in rectangular crops in one step. The 18,000,000 pixels are labeled as berry pixels or background pixels. In a similar way, about 1,800,000 labeled pixels are extracted that are labeled as wax pixels, non-wax pixels (i.e., pixels depicting berry surface without wax) or background pixels (i.e., pixels depicting background, lignifications or the pedicle of the berry). The extraction of labeled pixels for the evaluation of wax measurement is more difficult due to smaller crops showing the same classes of pixels. Each run of the 3-fold cross-validation of the CNN-based berry detection employs 12,000,000 labeled pixels taken from the 180 training images for training and 6,000,000 labeled pixels taken from the 90 test images for evaluation. As final result, we derive the averaged accuracy of the pixel classification over these three runs. Analogously, each run of the 3-fold cross-validation of the CNN-based wax measurement employs 1.200,000 labeled pixels for training and 600,000 labeled pixels for evaluation. Table \ref{table:table_1} shows the averaged accuracy values for berry detection and wax quantification as well as the averaged times for image capture.\\
The highest accuracies are obtained by using all inputs (i.e., input mode IV) showing improvements  of the accuracy values of 6.6\% and 8.1\% compared to the baseline mode I (only standard color images). The best accuracy values of a stand-alone separation method are obtained by using only the pattern-based light separation showing improvements of 4,9\% and 7,4\% compared to the baseline mode. Generally, we can observe that each of the light separation approaches improve the accuracy values for berry detection and wax quantification compared to the baseline mode. On the other hand, the time required by pattern-based light separation approach is a magnitude higher compared to the time requirement of the polarized-based light separation approach.
\begin{table}[H]
	\centering
		\begin{tabular}{c c c c}
			\toprule
			\textbf{Input Mode} & \makecell{\textbf{Accuracy} \\ \textbf{Berry} \\ \textbf{Detection}} & \makecell{\textbf{Accuracy} \\ \textbf{Wax} \\ \textbf{Detection}} & \makecell{\textbf{Capture} \\ \textbf{Time}}\\
			\midrule
			(I) Standard &  0,913 & 0,892 & $\sim$2-3 sec\\
			(II) Polarization-based separation & 0,970 &  0,945 & {$\sim$2-3 sec}\\
			(III) Pattern-based separation &  0,962 & 0,966 &{$ \sim$15 sec}\\
			\midrule
			(IV) Standard + pattern-based + polarization-based separation &  0,979 & 0,973 &$\sim$19-21 sec\\
			\bottomrule
		\end{tabular}
		\caption{Averaged accuracy values for berry detection and epicuticular waxes quantification as well as average time requirements for image capture.}
\label{table:table_1}
\end{table}

\subsection{Quantification of Epicuticular Waxes}
\label{subsection:subsection_2_2_Quantification_of_Epicuticular_Waxes}
\begin{figure}[H]
	\centering
	\includegraphics[scale=0.65]{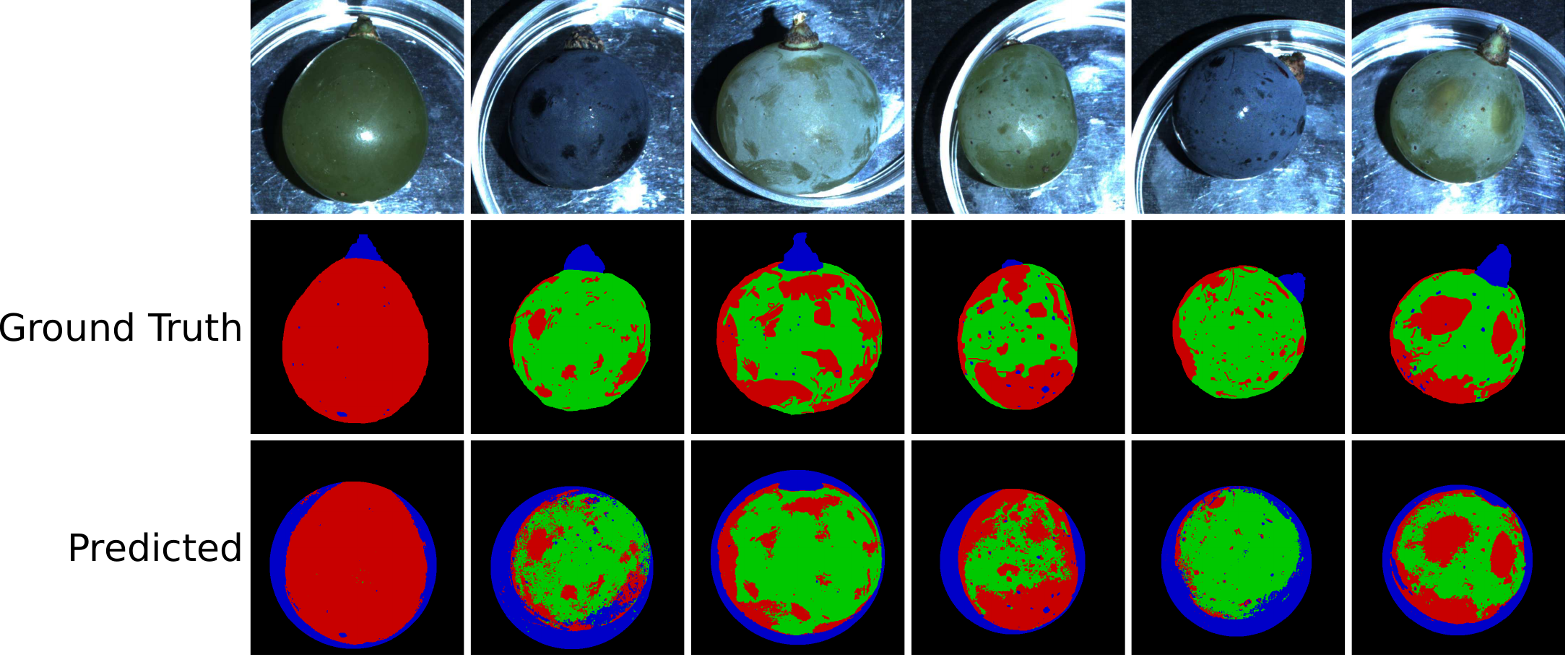}
	\caption{Quantification of epicuticular waxes (green) using the pattern-based light-separation method. Blue pixels depict background, lignification or pedicle within the area of interest. Red pixels depict regions of the berry surface without epicuticular waxes. Top row: standard color images. Middle row: Manually generated ground truth in order to visually compare with the estimated classification (Bottom row).These ground truths are not used to train the CNN.  From left to right: berries of the cultivars Morio Muskat, Dakapo, Seibel 7511, Sauvignon Blanc, Cabernet Sauvignon and Riesling.}
	\label{fig:figure_2}
\end{figure}

To characterize the uniformity and intactness of the epicuticular waxes, we derive a quantification of pixels depicting epicuticular waxes on a berry's surface. Figure \ref{fig:figure_2} shows the visualization of the areas covered by epicuticular waxes detection.

 As expected the cultivar Morio Muskat shows the lowest proportion of wax and thus, the lowest amount of berry bloom. The most consistent layer of epicuticular waxes was observed on berries of Cabernet Sauvignon, followed by Seibel 7511 with only small interruptions in the berry bloom.
 
\begin{figure}[H]
	\centering
	\includegraphics[scale=0.50]{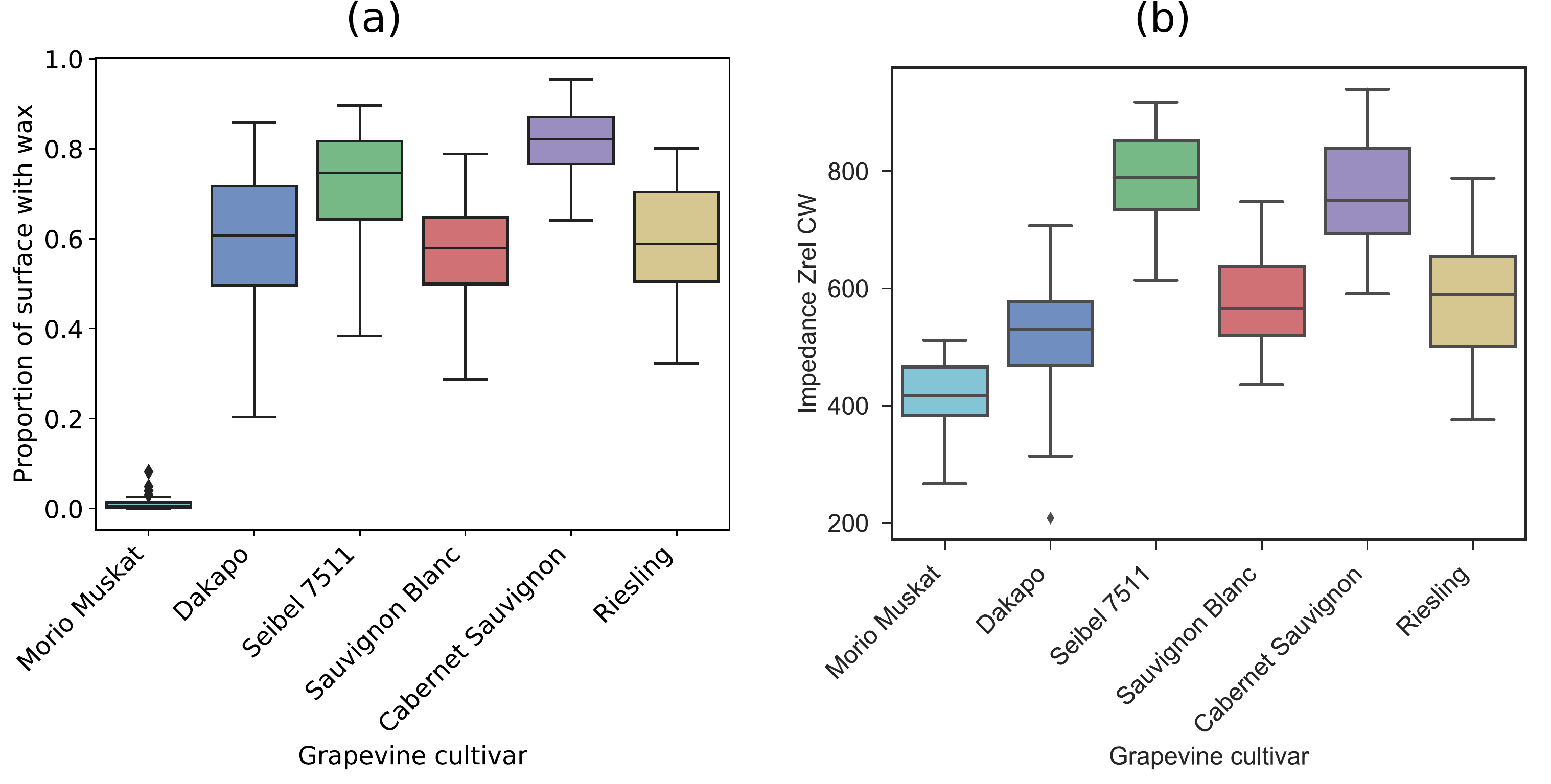}
	\caption{\textbf{(a)} Distribution of the proportion of epicuticular waxes on the berry surface per cultivar. \textbf{(b)} Distribution of the impedance $Z_{rel}CW$ on the berry surface per cultivar. With $N = 45$ berries for each cultivar, except for Cabernet Sauvignon in \textbf{(b)} where $N=44$ (see Supplementary Table S1).}
	\label{fig:figure_3}
\end{figure}

\begin{figure}[H]
	\centering
	\includegraphics[scale=0.70]{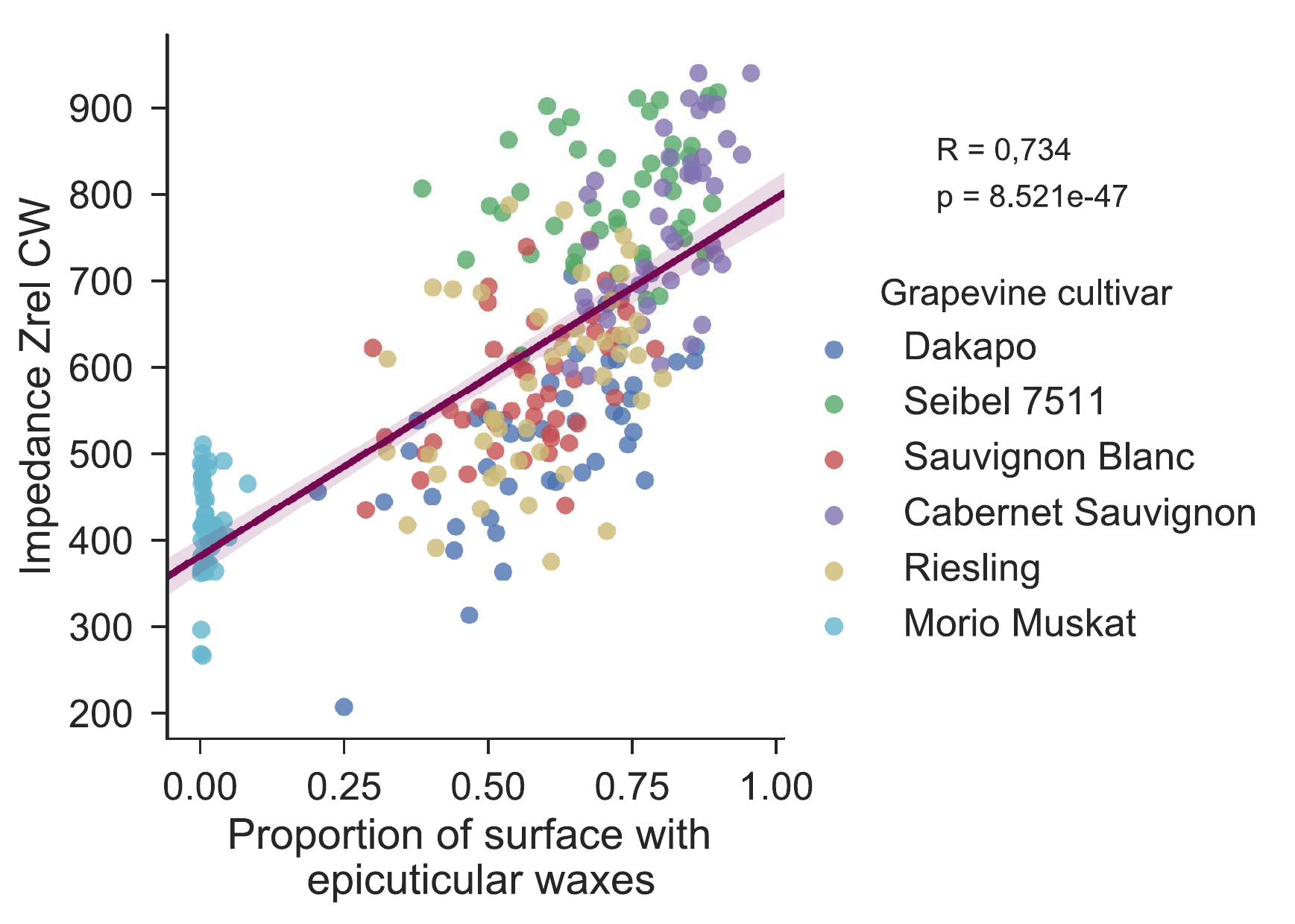}
	\caption{Correlation plot of the measured relative impedance $Z_{rel}CW$ and the  proportion of epicuticular waxes per cultivars using images of the combination of the three methods. N=269 investigated berries. Using the pattern-based method we obtain a correlation value of $R = 0,738$ with $\num{1.742e-47}$ and using the polarization-based method a correlation value of $R = 0.761$ with $\num{3,564e-52}$}
	\label{fig:figure_4}
\end{figure}

The left box plot in figure \ref{fig:figure_3} \textbf{(a)} depicts the distributions of the proportion of epicuticular waxes on the berry surfaces for each cultivar derived by the k-fold cross-validation with k = 3 on the sample of 270 berry images (cf. section \ref{subsection:section_2_1_Deep Learning_:_Detection_of_the_Berry_and_Its_Epicuticular_Waxes}) using the combination of all inputs, i.e., input mode IV. As visible in figure \ref{fig:figure_3} \textbf{(a)}, the detected proportion of the berry surface with epicuticular waxes partially differentiates considerable between the investigated cultivars. In contrast to the estimated OIV 227 classification (Table \ref{table:table_2}, section \ref{subsection:subsection_4_1_Data}) the highest proportion of epicuticular wax was obtained on berries of Cabernet Sauvignon, the proportion on berry surfaces of Riesling, Sauvignon Blanc and Dakapo was comparable. The variance of detected proportion of wax was larger for Dakapo and Riesling in comparison to Sauvignon blanc. Similar observations were obtained for the thickness of the cuticle, the impedance $Z_{rel}CW$. Both boxplots indicate a positive correlation between relative impedance values and proportions of epicuticular waxes. The relative impedance of the cuticle waxes is described as an indicator for the thickness and permeability of the cuticle and thus, is described as indicator for resilience towards \textit{Botrytis} bunch rot \cite{herzog_impedance_2015}. The right box plot in figure \ref{fig:figure_3} \textbf{(b)} depicts the distributions of the impedance values $Z_{rel}CW$ on the berry surfaces for each cultivar. Figure \ref{fig:figure_4} illustrates the relation between electrical impedance and the detected proportion of epicuticular waxes (computed with classification results using input mode IV) over the cultivars. The linear relation between the distribution of waxes and the electrical impedance indicates that both phenotypic traits (thickness and distribution/ uniformity) are linked.
\section{Discussion}
\label{section:section_3_Discussion}
The aim of the present study was the development of an automated and mobile light-separation camera system and the development of an automated CNN-based image analysis approach. 
We assumed that pattern-based and polarization-based light-separation methods are promising imaging tools in order to characterize the distribution and proportion of the grape berry epicuticular waxes. The best accuracy of the CNN for the berry detection could be observed with the application of the polarization-based method which needs only few seconds. Using this light-separation method for epicuticular wax detection, the accuracy decreased to 0.95 in comparison to the input mode IV (0.97). From the practical point-of-view, the accuracy of both is reliable enough for further investigations and phenotypic screenings. More important now is the time which is needed for single measurements because of the limited period of only a few days when the berry epicuticular wax can be phenotyped. The polarization-based method, however, seems to be the most promising manner in order to screen large samples e.g. within breeding programs.
Single light reflection components captured during light-separation process contain also important information about surface characteristics of grapevine berries especially about its epicuticular waxes, the berry bloom. This phenotypic information is successfully classified with the deep-learning method and so the proportion of epicuticular waxes are computed. We achieved positive correlations between the relative impedance and the proportion of epicuticular waxes. This correlation indicates the linear relation between the thickness and permeability of the cuticle (impedance) and the homogeneity of the wax distribution. Both characteristics are described as important traits to increase the resilience of cultivars and thus, the acquisition of both traits are promising to improve the selection procedure within grapevine breeding purposes regarding thickness and quality of berry bloom. The combination of both sensor-based approaches (visible and non-visible) will improve existing prediction models as described by \cite{herzog_impedance_2015} or \cite{molitor2018multi}. \\
We detected different proportions of waxes between the exemplary investigated cultivars. The intactness of the wax layer depends on the density of the bunches, i.e. normal bunches (OIV 204 =5) like the bunches of Cabernet Sauvignon, show a more consistent wax layer in comparison to cultivars like Riesling showing dense bunches. In dense bunches the berries come into contact with each other which interrupts the wax layer \cite{becker2012deposition}. Regarding the relation of a thick, consistent epicuticular wax and an increased resilience towards \textit{Botrytis} bunch rot, further investigations are needed. The application of the presented method enables non-invasive phenotyping of individual berries and subsequently, the usage of the same berries for Botrytis infection test under controlled lab conditions \cite{sapkota2015phenotypic}.  The system defines also a refined approach to measure quality and quantity of epicuticular waxes which is a good indicator for vulnerability to \textit{Botrytis} bunch rot. In addition, the application of the developed light-separation system could be extended due to the screening of genetic resources showing different wax compositions as shown by \cite{rid2018waxy}. The chemical analysis of such wax compositions is very time-consuming and expensive, it is not feasible to investigate wax content on single berry level thus the presented method may be a promising tool to detect different berry bloom characteristics due to different wax compositions. However, grapevine breeding purposes require (field) screenings of large breeding population, genetic repository or seedlings with more than 50,000 plants. The light-separation method is a promising tool for high-throughput applications in the field. For this, the first step would be to capture several berries at once and then part of wine grapes instead of a single berry per image, which requires to adapt berry and epicuticular waxes methods, applied a full-segmentation process such as by \cite{badrinarayanan2017segnet}. Through this new approach, images of whole grapes can be captured by enlarging the LSL to produce a sufficiently light scene over the whole grape. Secondly, the light-separation methods have to be adjusted to outdoor conditions, using for example approaches as described by \cite{garces_low_2015} or build a larger LSL that cover whole vines to neutralize the natural light. Finally, this field version of the LSL could be mounted on a mobile platform such as the PhenoBOT \cite{kicherer2015automated} or Phenoliner \cite{kicherer2017phenoliner} to obtain as much information as possible with light-separation-methods.

\section{Data and Methods}
\label{section:section_4_Data_and_Methods}
First, we will describe in more detail the generation of our sample set of 270 berry images captured by the LSL (cf. section \ref{subsection:subsection_4_1_Data}). Secondly, we will briefly explain the main principles of pattern and polarization-based light-separation (cf. section \ref{subsection:subsection_4_2_Light_Separation_Methods}). Thirdly, we will present the hardware and software aspects of the technical implementation of the LSL (cf. section \ref{subsection:subsection_4_3_LSL_and_I_Sensor} and section \ref{subsection:subsection_4_4_software}). Finally, we report relevant specifications of the architecture as well as of the training and evaluation settings of both convolutional neural networks for berry detection and epicuticular wax quantification (cf. \ref{subsection:subsection_4_5_CNN_Training}).

\subsection{Data}
\label{subsection:subsection_4_1_Data}
\begin{figure}[h]
	\centering
	\includegraphics[scale=0.9]{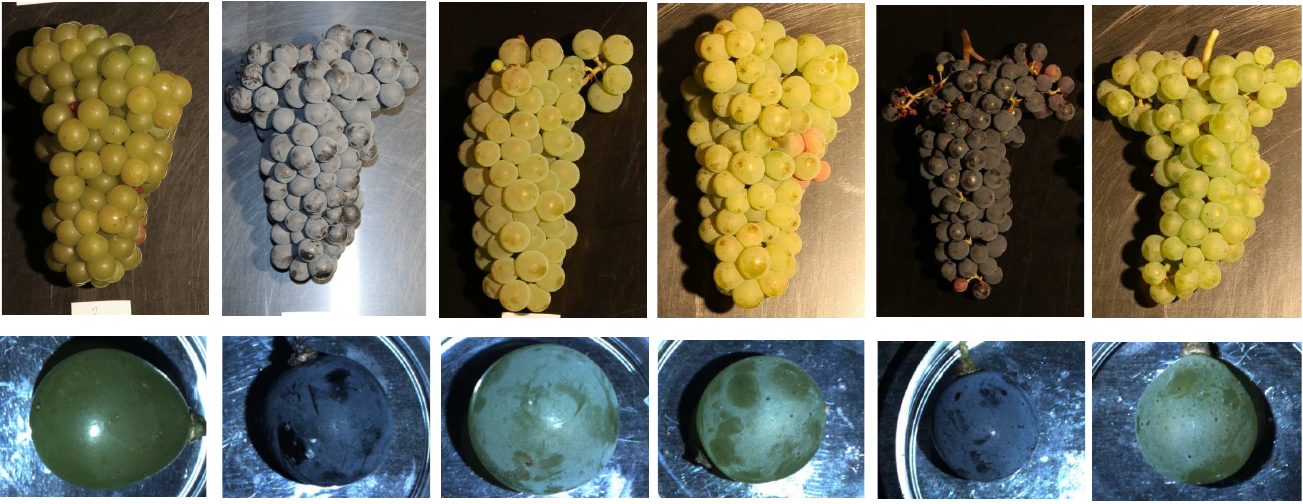}
	\caption{Overview of the six cultivars. From left to right and top to bottom: Morio Muskat, Dakapo, Seibel 	7511, Sauvignon blanc, Cabernet Sauvignon, Riesling.}
	\label{fig:figure_5}
\end{figure}

Our sample set of 270 berry images was captured using the mobile LSL at the Julius Kühn-Institute (JKI), Institute for Grapevine Breeding Geilweilerhof, Siebeldingen in Rhineland-Palatinate, Germany. The LSL-based capturing of the berry images was part of an overall measurement campaign at the JKI also including impedance measurements of the cuticle and its epicuticular waxes as additional objective phenotypic data. This measurement campaign included berries from six cultivars (also cf. figure \ref{fig:figure_5}), which are listed in table \ref{table:table_2}.
\begin{table}[H]
	\centering
	\begin{tabular}{c c c c}
		\toprule
		\textbf{Cultivar} & \textbf{OIV 204} & \textbf{OIV 227} & \makecell{\textbf{Susceptibility towards}\\\textbf{ \textit{Botrytis} bunch rot}} \\ 
		\midrule
		Morio Muskat & 9 & 1& highly susceptible (9)\\
		Dakapo & 7 & 5& medium susceptible (5)\\
		Seibel 7511 & 5 & 7& resilient (1)\\
		Sauvignon Blanc &7& 5&medium susceptible (5)\\
		Cabernet Sauvignon & 5& 5&resilient (1)\\
		Riesling& 7&3& susceptible (7)\\
		\bottomrule
	\end{tabular}
	\caption{The six cultivars and their respective descriptors OIV 204 (bunch density) and OIV 227 (berry bloom) (cf. \cite{oivClassification}) and the estimated susceptibility towards \textit{Botrytis} bunch rot.}
	\label{table:table_2}
\end{table}
Six different grapevine cultivars (table \ref{table:table_2} and figure \ref{fig:figure_5}) were selected due to their berry bloom (OIV 227) and their susceptibility towards \textit{Botrytis} bunch rot. All cultivars show compact bunch structure (OIV 204). Data acquisition was done in three tiers, each tier dealing with fifteen berries of each cultivar. Therefore, we acquired data from ninety berries per tier and 270 berries in total. All image samples were captured with LSL at the same day yielding for each berry one standard RGB image (baseline reference) and four light separated RGB images showing specular reflections, diffuse reflections, direct reflections, and global reflections, respectively (cf. section \ref{subsection:section_2_1_Deep Learning_:_Detection_of_the_Berry_and_Its_Epicuticular_Waxes} and table \ref{table:table_1}).

\subsection{Light Separation Methods}
\label{subsection:subsection_4_2_Light_Separation_Methods}
\begin{figure}[H]
	\centering
	\includegraphics[scale=0.6]{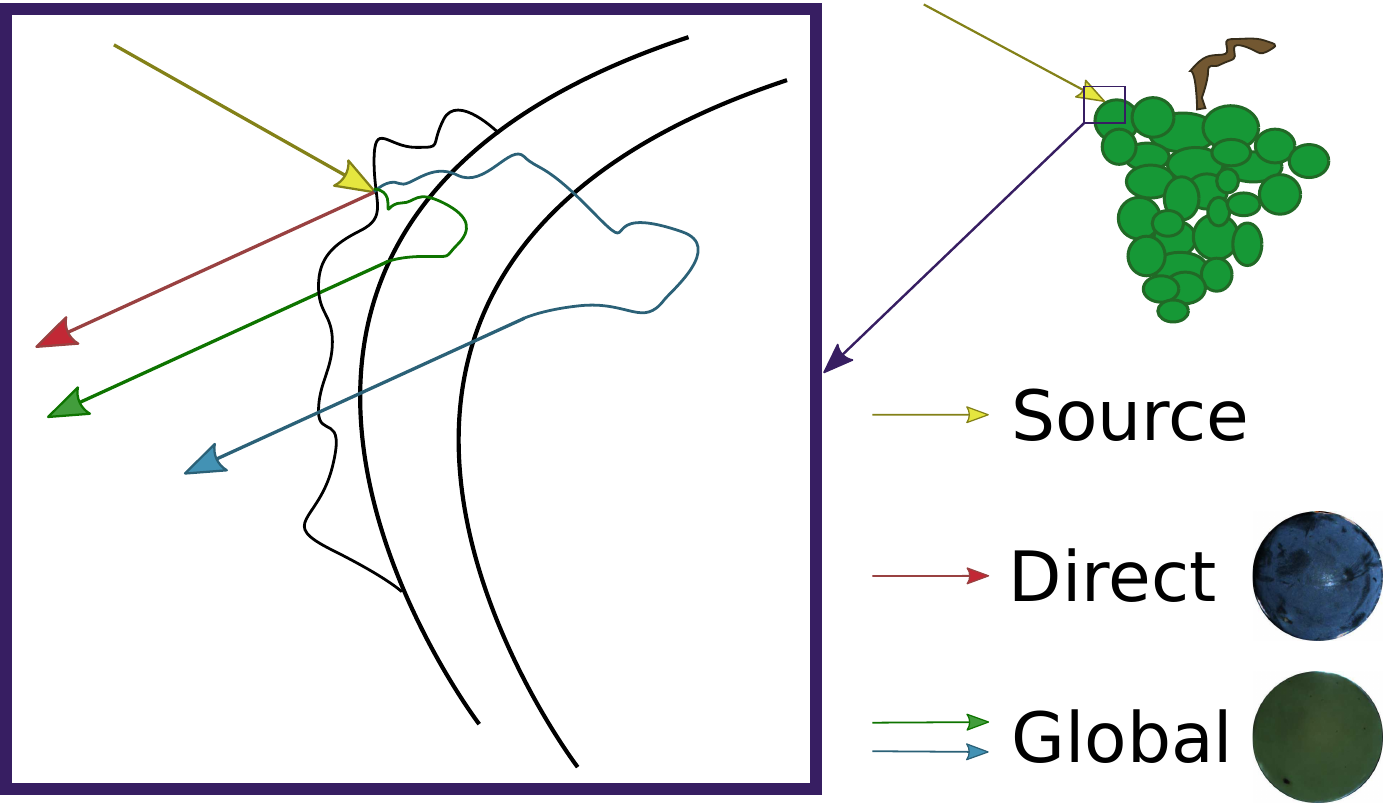}
	\caption{Direct and global illumination components. Measuring different illumination components of an object conveys information about the object's surface that cannot be inferred from their sum.}
	\label{fig:figure_6}
\end{figure}

The pattern-based method of Nayar et al. \cite{nayar_fast_2006} generates high frequency binary illumination patterns derived by chessboard-like arrangements of transparent and opaque filter cells positioned in front of the illumination source. While the light can pass the transparent filter cells it is blocked by the opaque filter cells. A light ray passing a transparent filter cell is part of the direct illumination by the source and will cause direct reflections on object surfaces. A patch of an object surface that is not illuminated directly from the light source (because it is shadowed due to an opaque filter cell) shows only global reflections caused by light rays that are reflected from other illuminated scene surfaces and by subsurface scattering within the medium beneath the object surface (cf. fig. \ref{fig:figure_6}). Of course, the directly illuminated surface patches receive also global illumination. Therefore, for all surface patches, the total radiance measured at a camera pixel is the sum of the direct and global components:
$I = I_{Global} + I_{Direct}$ .\\
In theory, two images of the scene taken with such a binary illumination pattern and its complementary binary pattern (transparent filter cells are now opaque transparent filter and vice versa) are sufficient to estimate the direct and global components for all surface patches (cf. equations \ref{eq:equation_1} and \ref{eq:equation_2} in section \ref{subsubsection:subsection_4_4_1_Data_Acquisition}). The direct reflections contain most surface information, i.e., about a berry's cuticle and its epicuticular waxes while the global reflections contain most subsurface information, like a berry's color appearance.

\begin{figure}[H]
	\centering
	\includegraphics[scale=0.55]{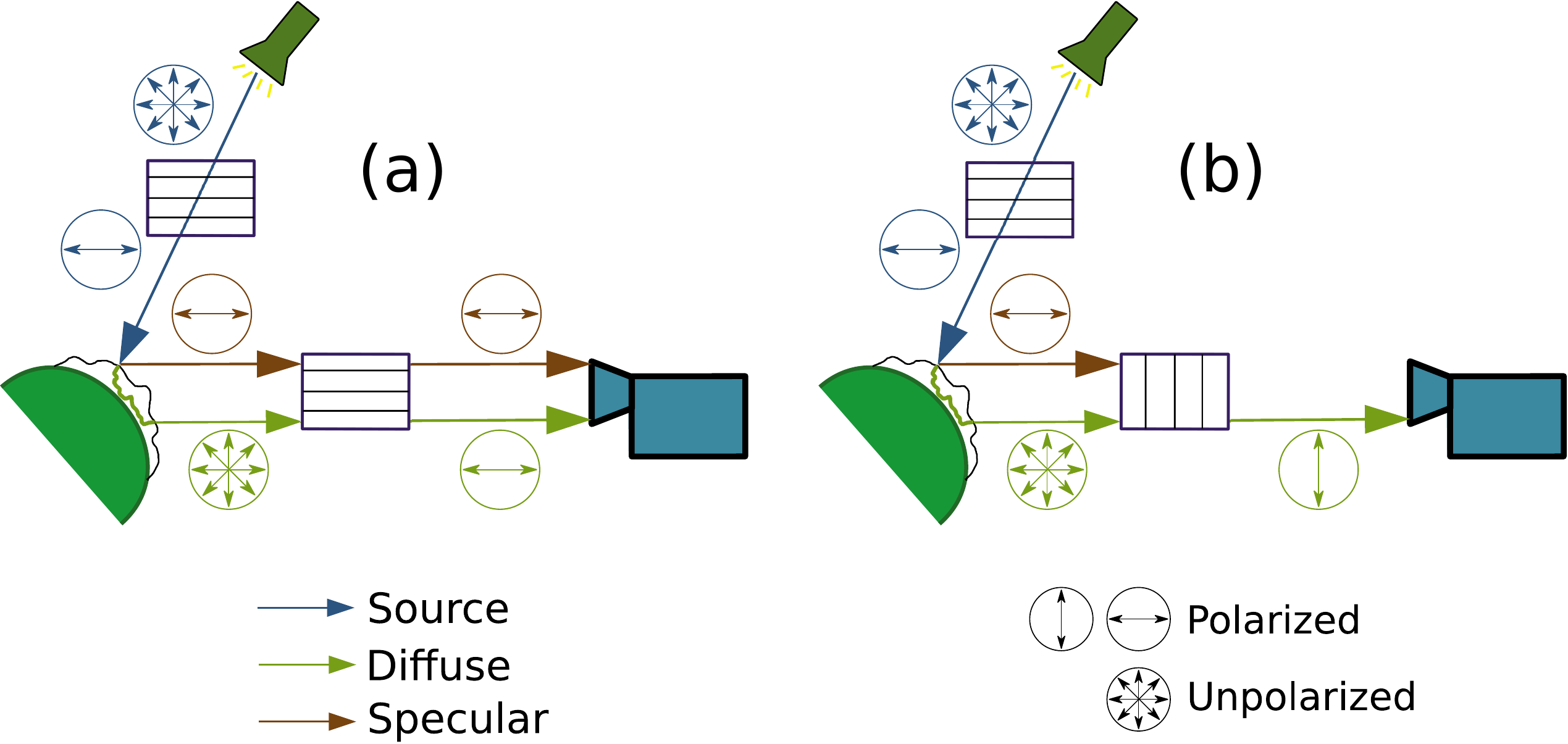}
	\caption{The polarization method is used to derive a parallel image and a perpendicular image. For both cases, the light from the illumination source is first polarized with linear polarized filter. This polarized light is then reflected in a unpolarized way from rough object surface patches (diffuse reflection) but it is reflected in a polarized way from smooth surface patches maintaining the polarization orientation (specular reflection). All reflected light is then polarized by a second filter at the front of the camera. \textbf{(a)} In the case of parallel oriented filters both types of reflections, i.e., diffuse and specular reflections are passed to the camera. \textbf{(b)} In the case of perpendicular oriented filters only the diffuse reflection can pass the filter to the camera.
	}
	\label{fig:figure_7}
\end{figure}

The polarization-method uses linear polarization filters to obtain two images called perpendicular image and parallel image. A linear polarization filter confines waves to only one plane. Parallel images are acquired by placing one linear polarization filter in front of the camera and a second one at the light source, where both filters show the same orientation (cf. fig. \ref{fig:figure_7} \textbf{(a)}). The perpendicular image in turn is derived by choosing perpendicular orientations of both filters (cf. fig. \ref{fig:figure_7} \textbf{(b)}). A pair of parallel and perpendicular images allows deriving the diffuse reflections and the specular reflections (cf. equations \ref{eq:equation_3} and \ref{eq:equation_4} in section \ref{subsubsection:subsection_4_4_1_Data_Acquisition}). Generally, specular reflections are caused by smooth surface patches (indicating no or little epicuticular waxes) while diffuse reflections are caused by rough surface patches (indicating epicuticular waxes).

\subsection{LSL (\textbf{L}ight \textbf{S}eparation \textbf{L}ab) and I-Sensor (Impedance Sensor)}
\label{subsection:subsection_4_3_LSL_and_I_Sensor}
\begin{figure}[h!]
	\centering
	\includegraphics[scale=0.20]{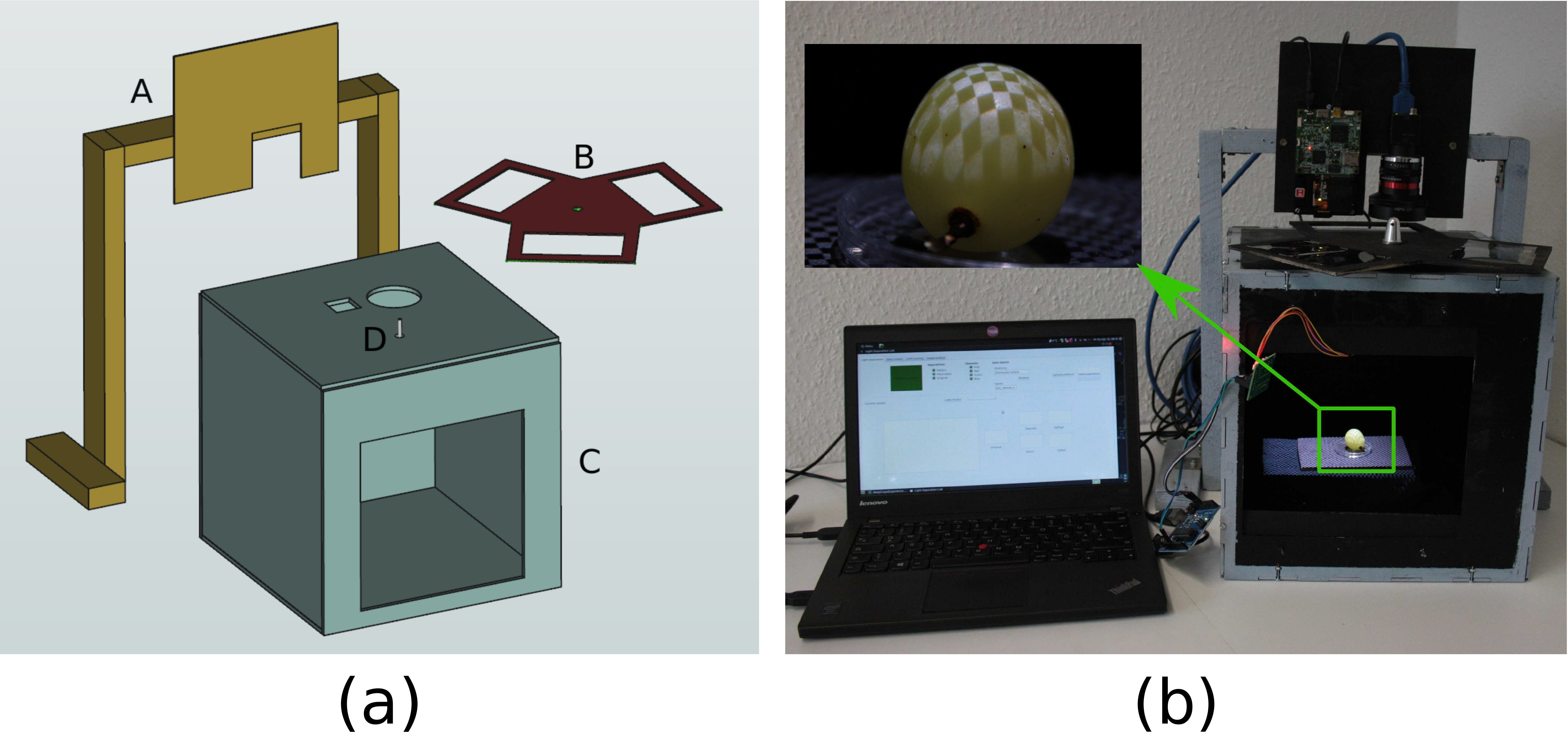}
	\caption{Main components of the LSL. \textbf{(a)} Technical sketch of the LSL with a scaffold for camera and beamer (A), the rotatable wheel for the polarization filters (B), the camera housing (C), and a screw fixed on the motor (D). \textbf{(b)} Constructed prototype of LSL controlled by a laptop. }
	\label{fig:figure_8}
\end{figure}

\begin{figure}[H]
	\centering
	\includegraphics[scale=0.35]{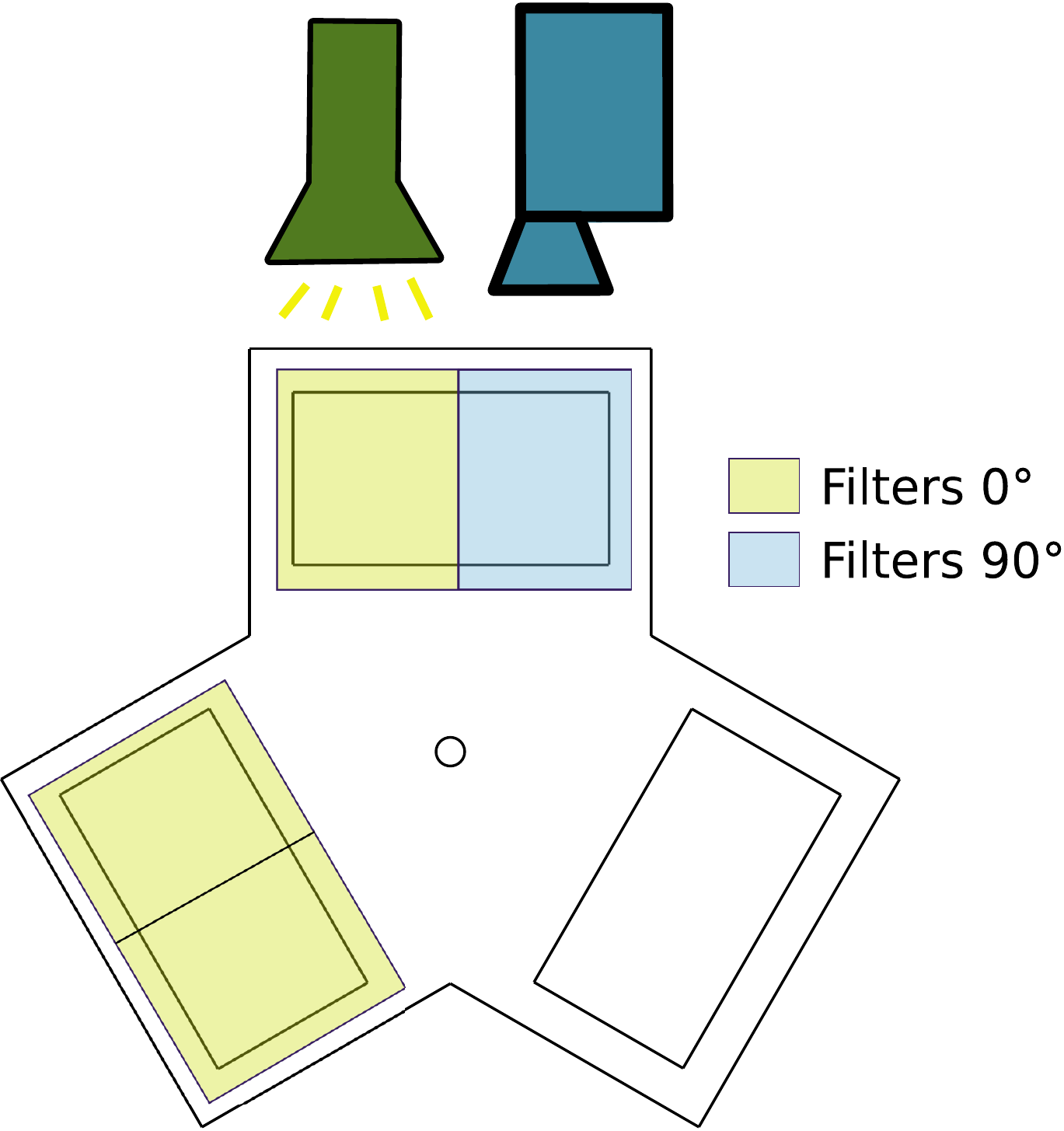}
	\caption{The wheel carrying the polarization filters shows three windows: one window without any linear polarization filter (for standard and pattern images), one window with parallel oriented filters and one window with two perpendicular oriented filters. Each window can be placed in front of camera and beamer.}
	\label{fig:figure_9}
\end{figure}

We designed the \textbf{L}ight \textbf{S}eparation \textbf{L}ab (LSL) to create a mobile solution for image capture including light-separation. The light in the LSL is generated by a DLP \textsuperscript{\textregistered} LightCrafter\texttrademark{} Evaluation Module from Texas Instruments and the camera used is a Flea3 (FL3-U3-88S2C-C) camera from FLIR. The Light Separation Lab shows four components (see figure \ref{fig:figure_8}):
\begin{itemize}
	\item a scaffold for camera and beamer (A)
	\item a wheel for the polarization filter (B)
	\item a camera housing (C)
	\item a stepper motor (D)
\end{itemize}
The scaffold positions camera and beamer next to each other. Thus, each of the three windows of the polarization wheel can be placed in front of both: camera and beamer. The polarization wheel is composed of three windows (cf. fig. \ref{fig:figure_9}): (1) one window without any filter (used when applying the pattern-based separation), (2) one window with parallel oriented filters (case (a) of fig. \ref{fig:figure_7}) and (3) one window with two perpendicular oriented filters (case (b) of fig. \ref{fig:figure_7}). For wheel positioning we use a stepper motor (28BY-J48) which is controlled by an Arduino Uno board and a ULN2003 driver board. \\
For an easy background separation we aimed for a background that reduces reflections and other interactions with illumination. First, we intended to use black velvet as background, but, since berries can lose juice while measuring, we just painted the interior of the LSL camera housing with black color. In the measurement campaign itself an additional small glass bin was used to position the berries quickly. For the next measurement campaign, we will provide a dark bin. Given the black background, berries of "white" grapes – that are actually green in color – are more easily distinguishable from the background. This is actually not the case for berries of "red" grapes – that are actually dark purple in color. \\
Impedance measurements were performed at room temperature using the "I-sensor" as developed by \cite{herzog_impedance_2015}. Relative impedance values Zrel were calculated according to \cite{herzog_impedance_2015}. The impedance measurements employ:
\begin{itemize}
	\item AD5933 high precision impedance converter system (Analog Devices GmbH,	Munich, Germany)
	\item USB-I2C module (Devantech Ltd (Robot Electronics, Norfolk, United Kingdom)
	using FTDI FT232R USB chip.
\end{itemize}

\subsection{Software}
\label{subsection:subsection_4_4_software}
\begin{figure}[h!]
	\centering
	\includegraphics[scale=0.44]{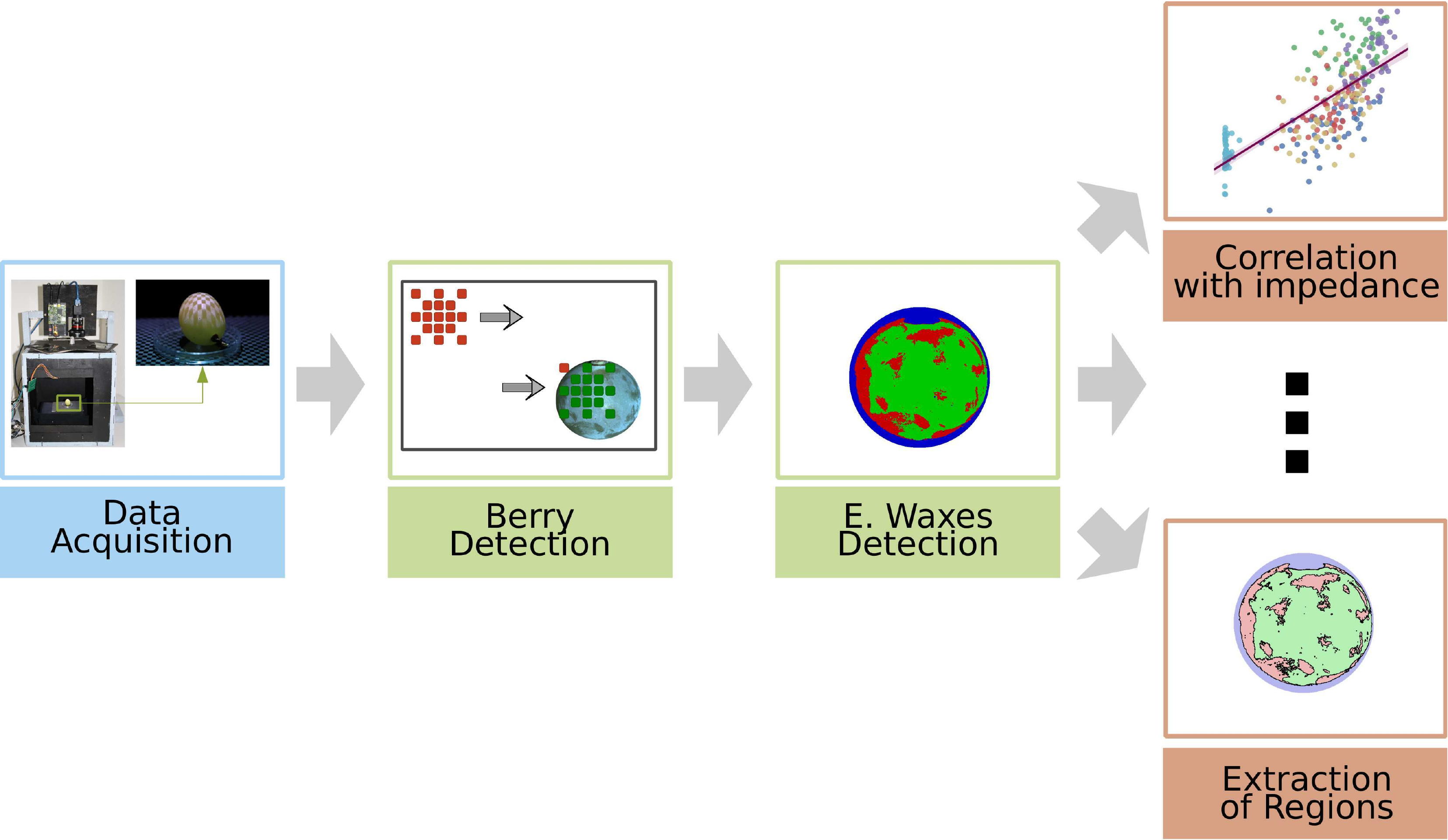}
	\caption{Workflow. First, standard, direct, global, diffuse and specular images of a berry are captured using the LSL. Secondly, berries are detected and localized in the images using a CNN-based shifting pattern recognition. Thirdly, epicuticular waxes are detected using a second CNN-based segmentation approach. Finally, the localization and quantification of the epicuticular waxes are used to analyze the proportion of waxes per berry or to extract coherent regions with wax or without wax etc. }
	\label{fig:figure_10}
\end{figure}
The first step of the process chain implements the data acquisition via the LSL. Secondly, two CNNs are used to localize the berry in the image and to detect epicuticular waxes on the berry’s surface. Finally, the detected epicuticular waxes are used for different analyses, for example to correlate impedance measure and proportion of epicuticular waxes or to extract single regions. Figure \ref{fig:figure_10} shows this workflow.

\subsubsection{Data Acquisition}
\label{subsubsection:subsection_4_4_1_Data_Acquisition}
To capture the light-separated images our controlling software uses the LightCrafter \cite{DlpApi}, FlyCapture \cite{flycaptureApi} and Arduino API \cite{arduinoApi} to control the hardware. Additionally we use the OpenCV \cite{bradski_opencv_2000} framework to transform the raw images delivered by the camera and save them in a datafile system. Before using the LSL, users can calibrate the hardware, e.g., initializing the position of the polarization wheel, or setting the focus of camera or/and beamer. Theoretically, the pattern-based method needs two images of chessboard-patterns in the best-case, each transparent and opaque filter cell having a size of $8 \times 8$ pixels. This size was derived empirically and turned out to capture the indirect reflection best. In practice, the boundaries of the squared filter cells are smooth due to the blurred projection of beamers. Therefore, we have to employ – according to Nayar et al. \cite{nayar_fast_2006} – a larger number of illumination patterns, so that each surface point of the scene shows maximum illumination as well as maximum darkness in different images. In our experiments, the same chessboard-pattern is twenty-five times shifted (five times vertically and five times horizontally) which ensures that each pixel of a scene image shows maximum and minimum brightnesses (cf. \cite{nayar_fast_2006}). Additionally, the black value of the beamer is computed, which represents the light intensity captured when the beamer projects only a black image. The twenty-five images derived that way are used to compute the direct and global images according to equations \ref{eq:equation_1} and \ref{eq:equation_2}, where the $b_{value}$ represents the computed black value (cf. \cite{nayar_fast_2006} and \cite{garces_low_2015}).

\begin{equation}
	I_{Direct} = min\{I_{Pat_{0}}, \dots, I_{Pat_{24}}\} - \frac{max\{I_{Pat_{0}}, \dots, I_{Pat_{24}}\}}{b_{value}-1}
	\label{eq:equation_1}
\end{equation}

\begin{equation}
	I_{Global} = 2 \times max\{I_{Pat_{0}}, \dots, I_{Pat_{24}}\} -\frac{I_{Direct}}{b_{value}+1}
	\label{eq:equation_2}
\end{equation}

The polarization method generates first a parallel image and a perpendicular image (cf. section \ref{subsection:subsection_4_2_Light_Separation_Methods}). The derivation of the two images showing the diffuse reflections and the specular reflections, respectively, follows equations \ref{eq:equation_3} and \ref{eq:equation_4}.

\begin{equation}
	I_{Diffuse} = 2 \times I_{Perpendicular}
	\label{eq:equation_3}
\end{equation}

\begin{equation}
	I_{Specular} = I_{Paralell} - \frac{I_{Diffuse}}{2}
	\label{eq:equation_4}
\end{equation}

\begin{figure}[h!]
	\centering
	\includegraphics[scale=0.6]{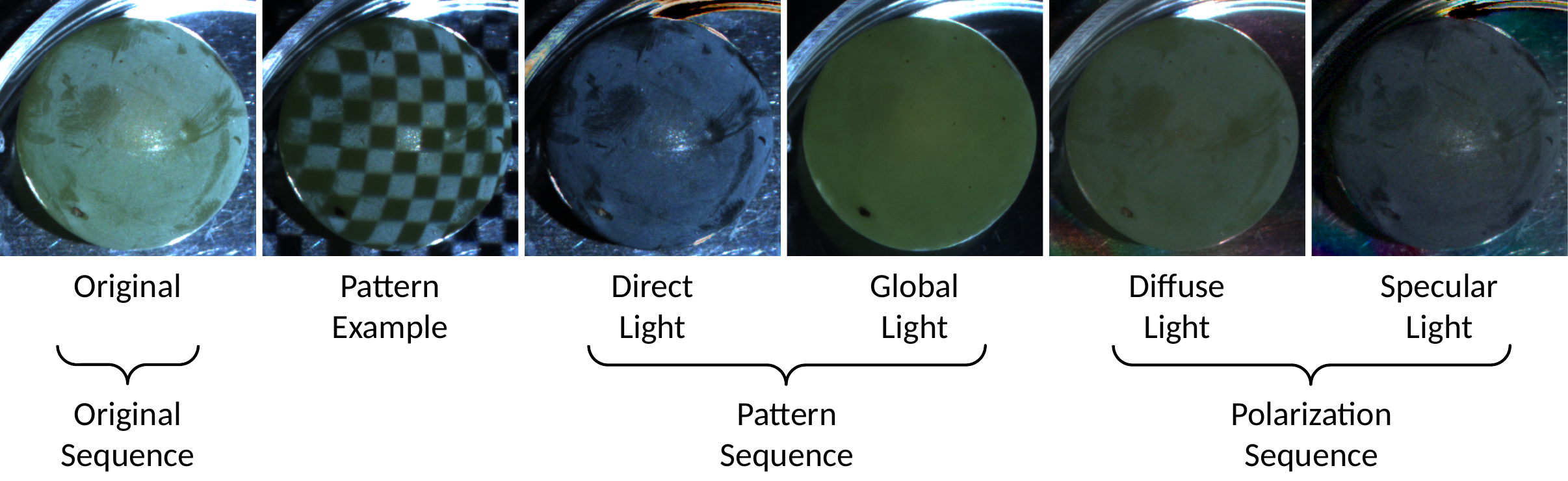}
	\caption{Overview of resulting images of the two light-separation processes, picture details, manually cropped ($400 \times 400$ pixels) out of images with dimensions $1920 \times 1080$ pixels. Berries are manually placed in the LSL, so we use our berry detector to localize them automatically in the image. We proceed with a gamma correction for both images of the polarization-based method in order to have a better visualization.}
	\label{fig:figure_11}
\end{figure}

Figure \ref{fig:figure_11} gives an impression of the image generation using both light separation methods. The first image is the standard RGB image generated without any light separation. The second image shows one example of an image illuminated by the projection of one of the 25 high frequency patterns. The third and fourth image depict the obtained direct and global reflection components while the last two images depict the diffuse and specular reflection components.

\subsubsection{Berry and Epicuticular Waxes Detection}
\label{subsubsection:subsection_4_4_2_Berry_and_Epicuticular_Waxes_Detection}
\begin{figure}[H]
	\centering
	\includegraphics[scale=0.85]{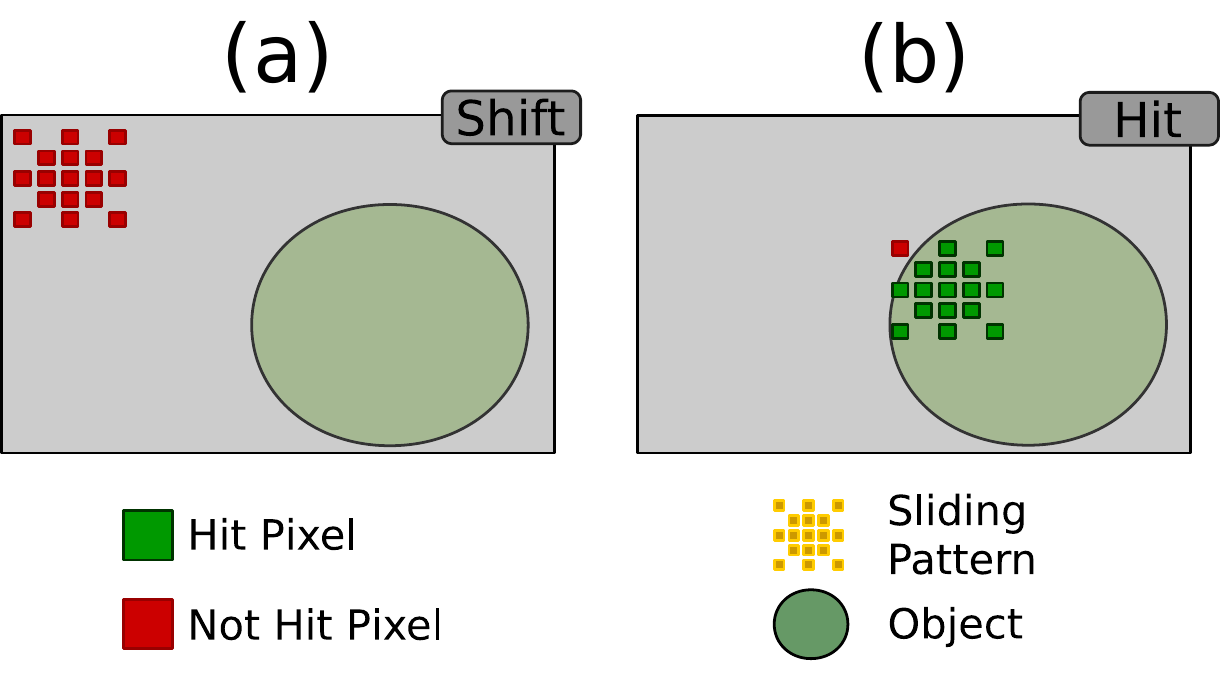}
	\caption{The sliding template method. \textbf{(a)} The 17 sensor pixels of the sliding pattern do not vote for berry detection. \textbf{(b)} The 17 sensor pixels of the sliding pattern vote for the detection of a berry. 
	}
	\label{fig:figure_12}
\end{figure}

We use two CNNs: one for the berry detection and one for the detection of the epicuticular waxes. A straightforward CNN-based approach to the berry detection would be to train and employ a CNN for pixel-wise classification, i.e. classifying each pixel of an input image into classes "berry" or "background". This approach shows two disadvantages: First, the annotation of the partially very small and arbitrary shaped waxed and non waxed surface areas (cf. fig. \ref{fig:figure_2}) is a very difficult, time consuming and error-prone work. Secondly, this approach would demand for the processing of complete full HD input images of size $1,920 \times 1,080 = 2,073,600$ pixels. But in the most cases, the berries are placed in the top-left part of the visual field. Therefore, runtime performance can be increased by using a sliding patch approach where a patch of size $128 \times 128 = 16,384$ pixels is scanning the full HD input image with an appropriate stepsize starting at the top-left corner of the input image. An additional speed-up can be achieved by using a sliding pattern template instead of a complete patch as input of a CNN. The pattern template shows only 17 selected pixels out of a complete patch. These 17 pixels form a pattern yielding a sparse representation of interior and boundary pixels of a patch (cf. \ref{fig:figure_12}). Given this pre-processing, there is no need to train a deep CNN with encoder and decoder part for a complete labeling of all pixels of patches. Instead, a by far smaller CNN is used on annotated image crops of size $3 \times 3$ pixels to learn the classification of the central pixel into the classes "berry" and "background".  Therefore, the CNN is very small with respect to the visual field and very efficient (cf. fig. \ref{fig:figure_13} a)). If the 17-pixel-pattern reports a hit, the centre and the radius of a spherical area of interest (AoI) as a first rough estimation of the visible berry shape can be determined by only four shifts of the pattern. After detection of the AoI, the second CNN is employed to classify all pixels in the AoI into the classes "epicuticular waxes", "no epicuticular waxes" and "others". Also this second CNN is designed as a small and efficient classification network that is trained again on annotated image crops of  size $3 \times 3$ pixels to learn the classification of the central pixel into the classes "epicuticular waxes", "no epicuticular waxes" and "others" (cf. fig. \ref{fig:figure_13} (b)). The annotation of training data and ground truth can now be done very effeciently using our semi-automated annotation tool that allows labeling all pixels in rectangular crops in one step. This annotation tool allows to label hundred of pixels in less than thirty seconds.\\
To cope with berries of different sizes, the pre-processing employs a multi-scale approach to the pattern-based berry detection. It starts with positioning the 17-pixel-pattern in patches of size $128 \times 128$ pixels. In the next scales the 17-pixel-pattern is positioned in patches of sizes $64 \times 64, 32 \times 32, 16 \times 16, 8 \times 8, 4 \times 4$, successively. The multi-scale iteration stops after having found the berry in one scale. Comparing this customized approach to a fully convolutional network approach (e.g., an AlexNet-based FCN with an up-convolution-based decoder part) working on the complete full HD images, we save more than 30 \% of the runtime in average.

\subsubsection{Results Analysis}
\label{subsubsection:subsection_4_4_3_Result_ Analysis}
Given the identification, localization and quantification of epicuticular waxes from the last CNN-based classification step, these results represent the starting point for various analyses and evaluations. In our current experiments for example, we evaluated the correlation between the visually derived quantification of epicuticular waxes with corresponding measurements of relative impedance values that we used for validation (cf. section \ref{subsection:subsection_2_2_Quantification_of_Epicuticular_Waxes} and the analysis step correlation with impedance in fig. \ref{fig:figure_10}). Another opportunity is the extraction and quantification of connected and coherent regions of epicuticular waxes (cf. the analysis step region extraction in fig. \ref{fig:figure_10}). The distributions and sizes of these regions of epicuticular waxes can give additional insights with respect to other phenotyping traits like compactness of grape bunches etc. We will consider additional options of analysis in our ongoing work. The visualization part of the result analysis is developed in Python 3.5  using the Numpy \cite{van_der_walt_numpy_2011}, Seaborn \cite{waskom_mwaskom/seaborn:_2017}, Matplotlib \cite{hunter_matplotlib:_2007} and Pandas \cite{mckinney_data_2010} libraries.

\subsection{CNN Training}
\label{subsection:subsection_4_5_CNN_Training}
\begin{figure}[h!]
	\centering
	\includegraphics{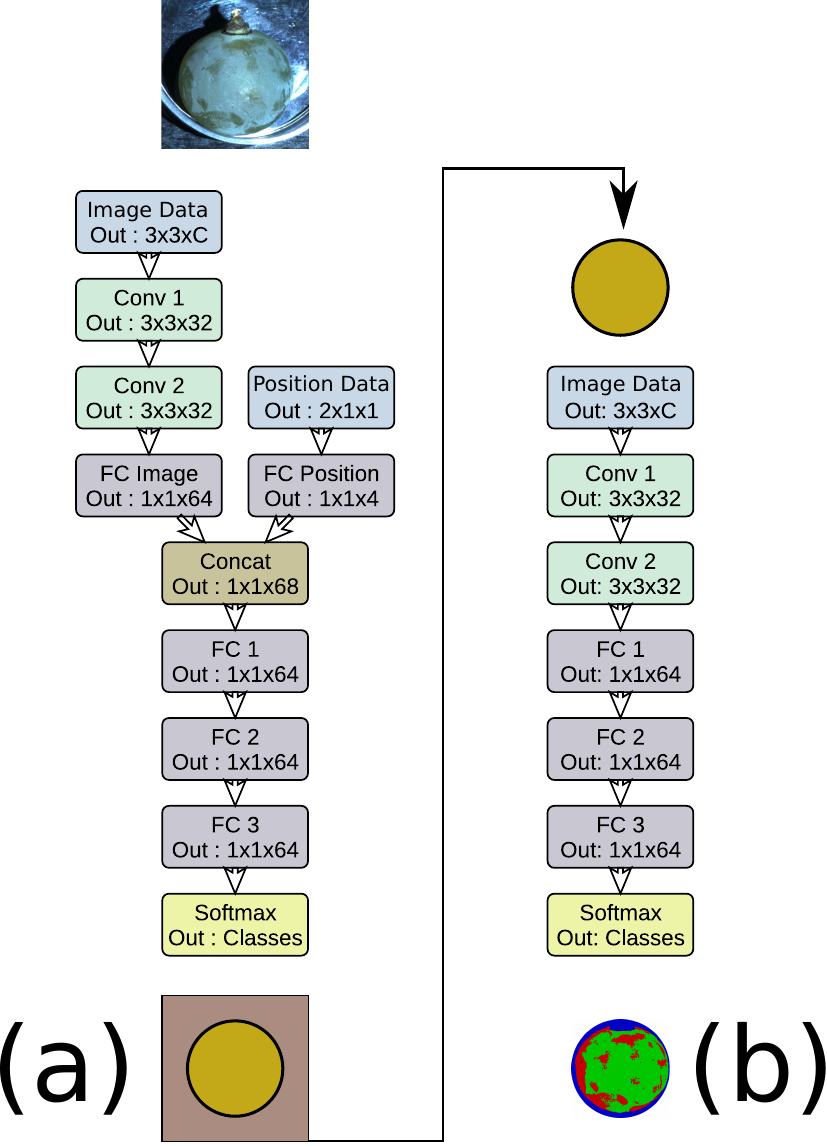}
	\caption{Both CNNs for respectively object and epicuticular waxes detection. There is a Rectified linear Unit (ReLU) after each of the fully-connected and convolution layers (\textbf{a}) Design of the CNN for the pixel-wise berry detection. (\textbf{b}) Design of the CNN for the pixel-wise epicuticular waxes.}
	\label{fig:figure_13}
\end{figure}

For the detection of berries and epicuticular waxes, we used two segmentation CNNs, i.e, each single pixel of the image input is classified into a category. The input for both CNNs is $3 \times 3$ crops of the image to learn the classification of the central pixel of a crop. We train both CNNs with the following parameters: momentum = 0.9, weight decay = $ 5\times 10^{-4}$ and a decreasing learning rate from $10^{-4}$ to $10^{-6}$ at training iterations 50.000 and 100.000, respectively. Training’s duration has been about thirty minutes for processing training datasets of about 6,000,000 and 600,000 pixels for berry detection and epicuticular waxes detection, respectively. The CNN for the berry detection processes also the pixel coordinates for the localization of the berries. The coordinates are represented in an additional input layer. A concatenation layer merges both resulting outputs (pixel values and pixel coordinates) as shown in figure \ref{fig:figure_13} \textbf{(a)}. After localization of the berry, the second CNN for wax detection is not in need for an additional input layer holding pixel coordinates (cf. figure \ref{fig:figure_13} \textbf{(b)}).

\section{Conclusions}
\label{section:section_5_Conclusions}
In order to phenotype the distribution of epicuticular waxes of the grapevine berry surfaces as important indicator for resilience towards \textit{Botrytis} bunch rot and as quality trait for table grape production, we 1) set up an automated and mobile light-separation camera system; and 2) developed an automated Convolutional Neural Network (CNN) based approach for fast image analysis. Regarding high-throughput applications in the future, especially with a view to breeders and the grape industry, several berries need to be phenotyped at once or on the go from an assembly line or in the field. Therefore, the most efficient light-separation setting needs to be determined and  a new design of the capturing box will be necessary, for example build an input/output mechanic with a conveyor belt to get berries under the camera and beamer,  so that the data acquisition process can be fully automated. Moreover, to take full advantage of the non-invasive quality of our approach, an outdoor adaption is conceivable by placing camera and beamer in a darkest possible field vehicle  
like the Phenoliner. The application of this non-invasive manner enables comparable monitoring studies and the precise characterization of large breeding material and genetic repositories as well as a promising quality control tool for table grape production. Hereby, automation, mobility, simple-to-apply and minimal user-interaction are important requirements in order to transfer the method to different labs or groups irrespective to scientific background. The availability of a user-friendly graphical user interface will extend the application field of the presented system.

\paragraph{Funding} We gratefully acknowledge the financial support of the Federal Office for Agriculture and Food (Bundesanstalt für Landwirtschaft und Ernährung, BLE) and the Federal Ministry of Food and Agriculture (Bundesministerium für Ernährung und Landwirtschaft, BMEL).This work was funded by BMEL in the framework of Vitismart (FZK 2815ERA05C) and the Deutsche Forschungsgemeinschaft (STE 806/2-1 and TO 152/6-1). Further, we thank the German Federal Ministry of Education and Research (Bundesministerium für Bildung und Forschung (BMBF), Bonn, Germany (NoViSys: FKZ 031A349E)).

\paragraph{Acknowledgments} We also thank Florian Schwander (JKI, Siebeldingen) for improving the setup of the I-Sensor and many thanks to Daniel Zendler (JKI, Siebeldingen) for the 3D-print of the I-Sensor casing. Further, we thank Jenny Mack (University of Bonn) for proofreading this article.

\paragraph{Abbreviations} The following abbreviations are used in this manuscript:\\

\noindent 
\begin{tabular}{@{}ll}
	CNN & Convolutional Neural Network \\
	LSL & Light Separation Labor \\
	RoI & Region of Interest \\
	ReLU & Rectified Linear Unit \\
\end{tabular}

\printbibliography

\end{document}